\documentclass{article}

\usepackage{microtype}
\usepackage{graphicx}
\usepackage{subfigure}
\usepackage{booktabs} 
\usepackage{enumitem}
\usepackage{amsmath,amssymb}
\DeclareMathOperator{\E}{\mathbb{E}}
\usepackage{multicol}
\usepackage{hyperref}

\usepackage{color}

\usepackage[accepted]{mlsys2020}

\mlsystitlerunning{Meta Federated Learning}

\begin{document}

\twocolumn[
\mlsystitle{Meta Federated Learning}







\begin{mlsysauthorlist}
\mlsysauthor{Omid Aramoon}{umd}
\mlsysauthor{Pin-Yu Chen}{ibm}
\mlsysauthor{Gang Qu}{umd}
\mlsysauthor{Yuan Tian}{uva}
\end{mlsysauthorlist}

\mlsysaffiliation{umd}{University of Maryland}
\mlsysaffiliation{ibm}{IBM Research}
\mlsysaffiliation{uva}{University of Virginia}

\mlsyscorrespondingauthor{Omid Aramoon}{oaramoon@umd.edu}

\mlsyskeywords{Federated Learning, Backdoor Attacks}

\vskip 0.3in

\begin{abstract}

Due to its distributed methodology alongside its privacy-preserving features, Federated Learning (FL) is vulnerable to training time adversarial attacks. In this study, our focus is on backdoor attacks in which the adversary's goal is to cause targeted misclassifications for inputs embedded with an adversarial trigger while maintaining an acceptable performance on the main learning task at hand.
Contemporary defenses against backdoor attacks in federated learning require direct access to each individual client's update which is not feasible in recent FL settings where \emph{Secure Aggregation} is deployed. In this study, we seek to answer the following question, \emph{”Is it possible to defend against backdoor attacks when secure aggregation is in place?”}, a question that has not been addressed by prior arts. To this end, we propose \emph{Meta Federated Learning} (Meta-FL), a novel variant of federated learning which not only is compatible with secure aggregation protocol but also facilitates defense against backdoor attacks.
We perform a systematic evaluation of Meta-FL on two classification datasets: SVHN \cite{SVHN} and GTSRB \citep{GTSRB}. The results show that Meta-FL not only achieves better utility than classic FL, but also enhances the performance of contemporary defenses in terms of robustness against adversarial attacks.
\end{abstract}
]



\printAffiliationsAndNotice{\mlsysEqualContribution} 

\section{Introduction}

Federated Learning (FL) is a distributed learning framework that enables millions of clients (e.g., mobile and edge devices) jointly train a deep learning model under the supervision of an orchestration server \citep{mcmahan2017communication,smith2017federated,zhao2018federated}. Taking advantage of training data distributed among the crowd of clients enables federated learning to train a highly accurate shared global model. Federated learning has gained significant interest from the industry with many tech companies, including Google and Apple deploying this framework to improve their services, such as next word prediction for messaging on mobile devices and voice recognition for digital assistants \citep{kairouz2019advances}.

In every round of federated learning, the central server randomly selects a cohort of participants to locally train the joint global model on their private data and submit an update to the server, which would be aggregated into the new global model. Federated learning decouples model training from the need to access participants' training data by collecting focused model updates that contain enough information for the server to improve the global model without revealing too much about the client's private data \citep{kairouz2019advances}. 

While collecting model updates, instead of centralizing raw training data, significantly reduces privacy concerns for participating clients, it does not offer any formal privacy guarantees. Recent studies have shown that model updates can still leak sensitive information about the client's data \citep{melis2019exploiting,nasr2018comprehensive}, which proves that preserving the privacy of clients is only a promise, and certainly not the reality of federated learning.

\begin{figure*}[t!]%
\centering
\subfigure[Meta federated learning]{%
\label{meta-fl-fig}%
\includegraphics[width=9cm,height=5cm]{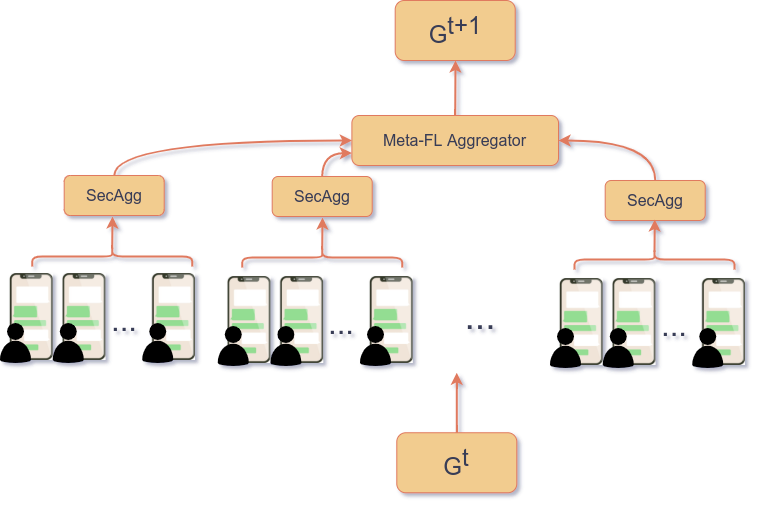}}%
\qquad
\subfigure[Federated learning]{%
\label{fl-fig}%
\includegraphics[width=3.84cm,height=5cm]{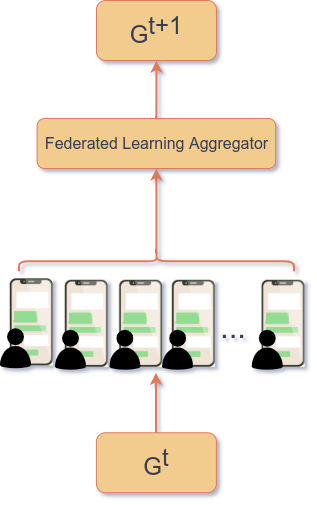}}%
\caption{Overview of model training in baseline and meta federated learning.}
\end{figure*}

To systematically address such privacy concerns, recent FL settings deploy Secure Aggregation (SecAgg) \citep{bonawitz2017practical}, a cryptographic protocol that enables server to compute aggregate of updates and train the global model while keeping each individual update uninspectable at all time. Looking from the server's point of view, secure aggregation can be a "double edged sword." On the one hand, it can systematically mitigate privacy risks for participants, which would make federated learning more appealing to clients and eventually result in higher client turnout. On the other hand, it would facilitate training time adversarial attacks by masking participants' contributions.

Training time adversarial attacks may have targeted \citep{chen2017targeted,liao2018backdoor,gu2019badnets} or untargeted adversarial objectives \citep{blanchard2017machine, mhamdi2018hidden}.
In untargeted attacks, the adversary aims to corrupt the learned model so that it'd perform poorly on the learning task at hand. However, in targeted attacks, adversary's goal is to force the model to learn certain adversarial sub-task in addition to the primary learning task. Targeted attacks are harder to detect compared to untargeted attacks for adversary's objective is unknown. Perhaps the most prevalent example of targeted attacks is backdoor attacks, which have been extensively explored for the centralized learning settings \citep{yao2019latent,li2019invisible,gu2019badnets}. In backdoor attacks, adversary's goal for the learned model is to misclassify inputs containing certain triggers while classifying inputs without the trigger correctly.

Known techniques in mitigating backdoor attacks in the centralized setting are not applicable to federated learning. Successful defenses such as data sanitization \citep{cretu2008casting} and network pruning \citep{liu2018fine} require careful examination of clients' training data or access to a proxy dataset with similar distribution as the global dataset. None of these requirements hold in federated learning.

Moreover, contemporary defenses against backdoor attacks in FL require examination of participant's model updates, which is not compatible with secure aggregation. Even in the absence of secure aggregation, inspecting client's update is not acceptable due to privacy concerns and regulations.

This paper seeks to answer the following question, \emph{" Is it possible to defend against backdoor attacks when secure aggregation is in place?"}, a question that has not been investigated by prior studies. To this end, we propose Meta Federated Learning (Meta-FL), a novel federated learning framework which not only preserves the privacy of participants but also facilitates defense against backdoor attacks.

In our framework, we take full advantage of the abundance of participants by engaging more than one training cohorts at each round to participate in model training. To preserve the privacy of participants, Meta-FL bootstraps the SecAgg protocol to aggregate updates from each training cohort. In Meta-FL, server is provided with a set of cohort aggregates, instead of individual model updates, which are further aggregated to generate the new global model. Figure \ref{meta-fl-fig} illustrates the overview of model training in Meta-FL.

Meta-FL moves defense execution point from update level to aggregate level which facilitates mitigating backdoor attacks by offering the following advantages: (i) server can monitor cohort aggregates without violating privacy of participants. Therefore, adversary is forced to be mindful of their submissions and maintain stealth on the aggregate level as aggregates which are statistically different from others are likely to get flagged and discarded; (ii) cohort aggregates exhibit less variation compared to individual client updates, which makes it easier for server to detect anomalies, and (iii) adversary faces competition from benign clients to hold control of the value of cohort aggregates which hinders them from executing intricate defense evasion techniques.

Our key contributions can be summarized as follows:
\begin{itemize}[leftmargin=*]
\setlength\itemsep{0em}
    \item We propose Meta federated learning, a novel federated learning framework that facilitates defense against backdoor attacks while protecting the privacy of participants.
    \item We show that moving the defense execution point from individual update level to aggregate level is effective in mitigating backdoor attacks without compromising privacy. 
    \item We perform a systematic evaluation of contemporary defenses against backdoor attack in both standard federated learning and Meta-FL. Results on two classification datasets: SVHN \cite{SVHN} and GTSRB \citep{GTSRB}, show that Meta-FL  enhances contemporary defense performance in terms of robustness to adversarial attacks and utility.
\end{itemize}

\section{Background}
\subsection{Federated Learning}
\label{baseline-FL-sec}
Federated learning is a machine learning setting that enables millions of clients (mobile or edge devices) to jointly train a deep learning model using their private data without compromising their privacy. The training procedure in federated learning is orchestrated by a central server responsible for providing the shared global model to participants and aggregating their submitted model updates to generate the new global model. The key appeal of federated learning is that it does not require centralizing participating users' training data, which makes it ideal for privacy-sensitive tasks.

A standard FL setting consists of P participating clients. Each client $i$ holds a shard of training data $D_{i}$ which is private to the client and is never shared with the orchestration server. In each round $t$ of federated learning, the central server randomly selects a set $\zeta^{t}$ of $c$ clients, and broadcasts the current global model $G^{t}$ to them. Selected set of clients $\zeta_{t}$ is referred to as \emph{training cohort} of round $t$. Each client $i$ in the training cohort locally and independently trains the joint model $G^{t}$ using Stochastic Gradient Descent (SGD) optimization algorithm for $E$ epochs on its local training data $D_{i}$ to obtain a new local model $L^{t+1}_{i}$, and submits the difference $L^{t+1}_{i} -G^{t}$ as its model update to the central server. Next, the central server averages model updates submitted by clients in the training cohort and updates the shared global model using its learning rate $\eta$ to obtain the new global model $G^{t+1}$, as shown in Equation \ref{FedAvg-Eq}. Model training resumes until the global model converges to an acceptable performance, or certain training rounds are completed. 
\begin{equation}
\label{FedAvg-Eq}
    G^{t+1} = G^{t}+\frac{\eta}{n}\sum_{i=i}^{n}(L^{t+1}_{i} -G^{t})
\end{equation}
\subsection{Secure Aggregation}
\label{sec-agg-sec}
Secure Aggregation (SecAgg) \citep{bonawitz2017practical} is a secure multi-party computation protocol that can reveal the sum of submitted model updates to the server (or aggregator) while keeping each individual update uninspectable at all time. Secure Aggregation consists of three phases, \emph{preparation}, \emph{commitment} and \emph{finalization} \citep{bonawitz2019towards}. In the preparation phase, shared secrets are established between the central server and participating clients. Model update from clients who drop out during the preparation phase will not be included in the aggregate. Next, in the commitment phase, each device uploads a cryptographically masked model update to the server, and the server computes the sum of the submitted mask updates. Only clients that successfully commit their masked model updates will contribute to the final aggregate. Lastly, in the finalization phase, committed clients reveal sufficient cryptographic secrets to allow the server to unmask the aggregated model update. 
\subsection{Robust Aggregation Rules and Defenses}
\label{Byz-Rob-Agg-Rules-sec}
Numerous studies have proposed robust aggregation rules \citep{blanchard2017machine,yin2018byzantine,pillutla2019robust} to ensure convergence of distributed learning algorithms in the presence of adversarial actors. The majority of studies in this line of work assume a byzantine threat model in which the adversary can cause local learning procedures to submit any arbitrary update to ensure convergence of learning algorithms to an ineffective model.
In addition to robust aggregation rules, several works have proposed novel defenses \citep{fung2018mitigating,sun2019can} against backdoor and poisoning attacks in federated learning. In what follows, we review several of the techniques which we experiment in Section \ref{experiment-sec}.

\textbf{Krum. }
The Krum algorithm, proposed by \cite{blanchard2017machine}, is a robust aggregation rule which can tolerate $f$ byzantine attackers out of $n$ participants selected at any training round.  
Krum has theoretical guarantees for the convergence should the condition $n \geq 2f + 3$ hold true. At any training round, for each model update $\delta_{i}$, Krum takes the following steps: (a) computes the pairwise euclidean distance of $n-f-2$ updates that are closest to $\delta_{i}$, (b) computes the sum of squared distances between update $\delta_{i}$ and its closest $n-f-2$ updates. Then, Krum chooses the model update with the lowest sum to update the parameters of the joint global model.   

\textbf{Coordinate-Wise Median. }
In Coordinate-Wise Median (CWM) aggregation rule \citep{yin2018byzantine}, for each $j_{th}$ model parameter, the $j_{th}$ coordinate of received model updates are sorted, and their median is used to update the corresponding parameter of the global model.

\textbf{Trimmed Mean. }
Trimmed Mean (TM) is a coordinate wise aggregation rule \citep{yin2018byzantine}. for $\beta \in [0,\frac{1}{2})$, trimmed mean computes the $j_{th}$ coordinate of aggregate of $n$ model updates as follows: (a) it sorts the $j_{th}$ coordinate of the $n$ updates, (b) discards the largest and smallest $\beta$ fraction of the sorted updates, and (c) takes the average of remaining $n(1-2\beta)$ updates as the aggregate for the $j_{th}$ coordinate.   

\textbf{RFA. }
RFA \citep{pillutla2019robust} is a robust privacy-preserving aggregator which requires a secure averaging oracle. RFA aggregates local models by computing an approximate of the geometric median of their parameters using a variant of the smoothed version of Weiszfeld’s algorithm \citep{weiszfeld1937point}. RFA appears to be tolerant to data poisoning attacks but can not offer byzantine tolerance as it still requires clients to compute aggregation weights according to the protocol. Relying on clients to follow a defensive protocol without a proper means to attest to the correctness of computations on the client-side cast doubts on the practicality of RFA. To the best of our knowledge, RFA is the only existing defense which is compatible with secure aggregation.

\textbf{Norm Bounding. }
Norm Bounding (NB) is an aggregation rule proposed by \cite{sun2019can}, which appears to be robust against false-label backdoor attacks. In this aggregation rule, a norm constraint $M$ is set for model updates submitted by clients to normalize the contribution of any individual participants. Norm bounding aggregates model updates as follows: (a) model updates with norms larger than the set threshold $M$ are projected to the $l2$ ball of size $M$ and then (b) all model updates are averaged to update the joint global model.     

\textbf{Differential Privacy. }
Differential Privacy (DP) originally was designed to establish a strong privacy guarantee for algorithms on aggregate databases, but it can also provide a defense against poisoning attacks \citep{ma2019data,dwork2006calibrating}. Extending DP to federated learning ensures that any participant's contribution is bounded and therefore, the joint global model does not over-fit to any individual update. DP is applied in FL as follows \citep{kairouz2019advances}: (a) server clips clients' model update by a norm $M$, (b) clipped updates are aggregated, then (c) a Gaussian noise is added to the resulted aggregate. DP has recently been explored and shown to be successful against false-label backdoor attacks in a study by \cite{sun2019can}.

\section{Problem Definition and Threat Model}
\label{problem-def-threat-model-sec}
In this section, we present the objectives, capabilities, and schemes of backdoor attackers that are commonly used in the prior studies. In other words, our proposed Meta-FL framework does not make any additional assumptions. 
\textbf{Attacker's Objective.} Similar to prior arts such as \citep{Xie2020DBA,bagdasaryan2020backdoor}, we consider an adversary whose goal is to cause misclassifications to a targeted label $T$ for inputs embedded with an attacker-chosen trigger. As opposed to Byzantine attacks \citep{blanchard2017machine}, whose purpose is to convergence the learning algorithm to a sub-optimal or utterly ineffective model, the adversary's  goal in backdoor attacks is to ensure that the joint global model achieves high accuracy on both the backdoor sub-task and the primary learning task at hand.

\textbf{Attacker's Capability.} 
We make the following assumptions about attacker's capabilities: (a) We assume attacker controls a number of participants, which are referred to as \textit{sybils} in the literature of distributed learning. Sybils are either malicious clients which are injected into federated learning system or benign clients whose FL training software has been compromised by the adversary, (b) following Kerckhoffs's theory \citep{shannon1949communication}, we assume a strong attacker who has complete control over local data and training procedure of all its sybils. The attacker can modify training procedure's hyperparameters and is capable of modifying model updates before submitting them to the central server, (c) adversary is not capable of compromising the central server or influencing other benign clients, and more importantly, does not have access to benign clients' local model, training data and submitted updates.

\textbf{Attack scheme.}
In our evaluations, we consider two backdoor attack schemes which are referred to as "Naive" and "Model Replacement" in literature \citep{bagdasaryan2020backdoor}. In both schemes, adversaries train their local model with a mixture of clean and backdoored data, and model updates are computed as the difference in the parameters of the backdoored local model and the shared global model. In the naive approach, the adversary submits the computed model update. While in model replacement attack, the model update is scaled using a scaling factor to cancel the contribution of other benign clients and increase impact of the adversarial update on the joint global model. A carefully chosen scaling factor for adversarial updates can guarantee the replacement of the joint global model with adversary's backdoored local model.

\section{Meta Federated Learning}
\label{meta-fl-sec}
In this section, we first discuss the challenges in mitigating backdoor attacks in federated learning. Then, we propose Meta Federated Learning (Meta-FL), and explain how it improves robustness to backdoor attacks while preserving the privacy of participating clients.

Challenges in defending against backdoor attack in federated learning are two-fold:

\textbf{Challenge 1.} Inspecting model updates is off limits with or without secure aggregation. Recent studies have demonstrated that model updates can be used to partially reconstruct clients' training data \citep{yao2019latent,li2019invisible,gu2019badnets}; therefore, any defensive approach which requires examination of submitted updates is a threat to privacy of participants, and against privacy promises of federated learning. Moreover, inspecting model updates simply is not be a valid option in systems augmented with SecAgg. Privacy promises of federated learning prohibits server from auditing clients' submissions which gives the adversary the privilege to submit any arbitrary value without getting flagged as anomalous. We refer to this privilege as \textbf{\textit{submission with no consequences}}.

\textbf{Challenge 2.} Even without restrictions mentioned above, defending against backdoor attacks would not be a trivial task. Model updates submitted by clients show high variations which makes it extremely difficult for the central server to identify whether an update works toward an adversarial goal. Sporadicity observed from model updates originates from the non-i.i.d distribution of original dataset among participants, and the fact that each update is product of stochastic gradient descent, a non-deterministic algorithm whose output is not merely a function of its input data.

Motivated to address the challenges above, we propose, Meta-FL, a novel federated setting which not only protects privacy of participants but also aids server in defending against backdoor attacks. Algorithm \ref{Meta-fl-alg} summarizes different steps of model training in our framework, which we will cover in detail here. 

In each round $t$ of training in Meta-FL, central server randomly selects {\large $\pi$} cohorts $\{\zeta_{1}^{t},\zeta_{2}^{t},..\zeta_{\pi}^{t}\}$, each containing $c$ unique clients (Line 3). Training cohorts can be sampled in-order or independently. In the latter case, each cohort is sampled after another, and thus, no client will be a member of more than one cohort ($\zeta_{i}^{t} \cap \zeta_{j}^{t} = \emptyset$). In the recent case, there is no inter-dependency among cohort selection, and therefore, cohorts can have clients in common; this scenario is more suitable for cases where ($P \leq$ {\large$\pi c$}). Next, server broadcasts global model $G^{t}$ to clients in each cohort (Line 5), each client $i$ locally and independently trains the model $G^{t}$ on their local training data to obtain a new local model $L^{t+1}_{i}$ , and compute their model update {\large$\delta_{i}$} as $L^{t+1}_{i} -G^{t}$ (Line 7).
Then, server establishes {\large $\pi$} separate instances of SecAgg protocol to concurrently compute aggregate of updates submitted from clients of each cohort (Line 9). 
Finally, in the last stage of training in Meta-FL, central server aggregates the "cohort updates" using aggregation rule $\Gamma$, and updates the joint model with its learning rate {\large$\eta$} to obtain next shared global model $G^{t+1}$, as shown in Line 11 of Algorithm \ref{Meta-fl-alg}. 

\begin{algorithm}[t]
\caption{Meta-FL framework}
\label{Meta-fl-alg}
\begin{algorithmic}[1]
\STATE Initialize shared global model
\FOR {each round t in 1,2,3..}
    \STATE Select $\pi$ training cohorts $\{\zeta_{1}^{t},\zeta_{2}^{t},..\zeta_{\pi}^{t}\}$ with $|\zeta_{1}^{i}|=c \,\, \forall i \in \{1,2,..,\pi\}$.
    \FOR{cohort $\zeta_{j}^{t}$ in  $\{\zeta_{1}^{t},\zeta_{2}^{t},..\zeta_{\pi}^{t}\}$ \textbf{in parallel}}
        \STATE Broadcast global model $G^{t}$ to cohort members.
        \FOR{client $i$ in  cohort $\zeta_{j}^{t}$ \textbf{in parallel}}
            \STATE $\delta^{t}_{i} \leftarrow ClientUpdate (i,G^{t})$
        \ENDFOR
        \STATE $\Delta^{t}_{j} \leftarrow SecAgg(\delta^{t}_{1},\delta^{t}_{2},..,\delta^{t}_{c})$
    \ENDFOR
\STATE $G^{t+1} = G^{t} + \eta \,  \Gamma(\Delta^{t}_{1},\Delta^{t}_{2},..,\Delta^{t}_{\pi})$  
\ENDFOR
\end{algorithmic}
\end{algorithm}

In our framework, plain model updates never leave client's side. All participants are required to follow the SecAgg protocol and submit cryptography masked updates. SecAgg guarantees that server is able to aggregate the masked submissions to update the global model but can not obtain value of each individual update. While each cohort aggregate may still leak information about collective training data of cohort members, the inferred information can not be associated to any individual client; therefore, privacy of participants is preserved in Meta-FL.  

In Meta-FL, as central server can only see aggregate of training cohorts, defense mechanisms are obliged to carry out on aggregate level rather update level. This property offers server several advantages in mitigating backdoor attacks, which we will cover in rest of this section. However, before we can proceed, we need to define several concepts that are key in understanding of what follows.

In rest of this paper, we refer to a training cohort as adversarial if and only if there exist at least one malicious client among its members. Naturally, a cohort is referred to as benign if none of its member are malicious. Moreover, we refer to aggregate of updates from a benign and an adversarial cohort as a benign and adversarial aggregate, respectively.

Moving defense execution point from update level to aggregate level facilitates mitigating backdoor attacks as it offers the following advantages:

\textbf{Advantage 1.} Server is allowed to inspect and monitor cohort aggregates. This property forces the adversary to maintain stealth on aggregate level as adversarial aggregates which are statistically different from other benign aggregates are likely to get detected and discarded by the server. Therefore, Meta-FL, revokes the privilege of submission with no consequences for adversary.

\textbf{Advantage 2.} Cohort aggregates are less sporadic compared to individual client updates which aids server in detecting anomalies. This advantage takes on the \emph{challenge 2} discussed above. By drawing an analogy to \emph{simple random sampling} in statistics \citep{rice2006mathematical}, we demonstrate that cohort aggregates show less variation across each coordinate compared to individual updates.

\begin{figure*}[h]%
\centering
\subfigure[SVHN]{%
\includegraphics[width=8cm,height=6cm]{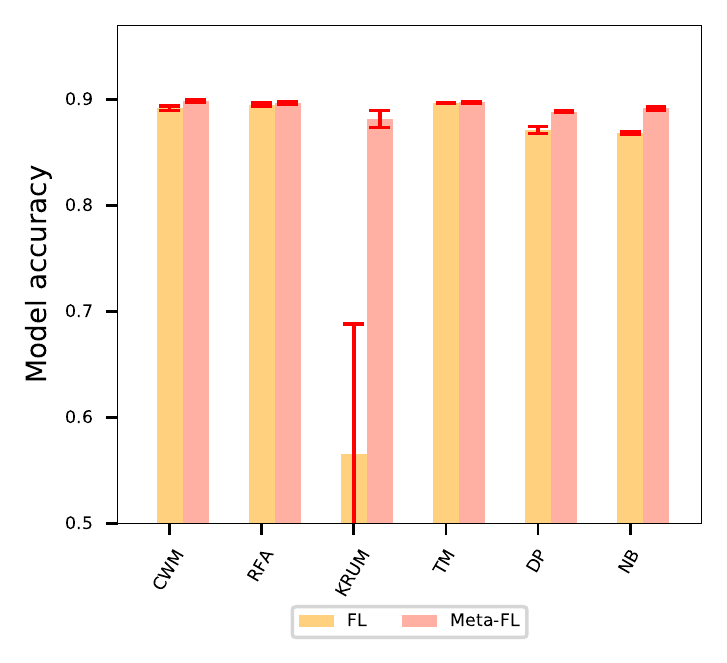}}%
\subfigure[GTSRB]{%
\includegraphics[width=8cm,height=6cm]{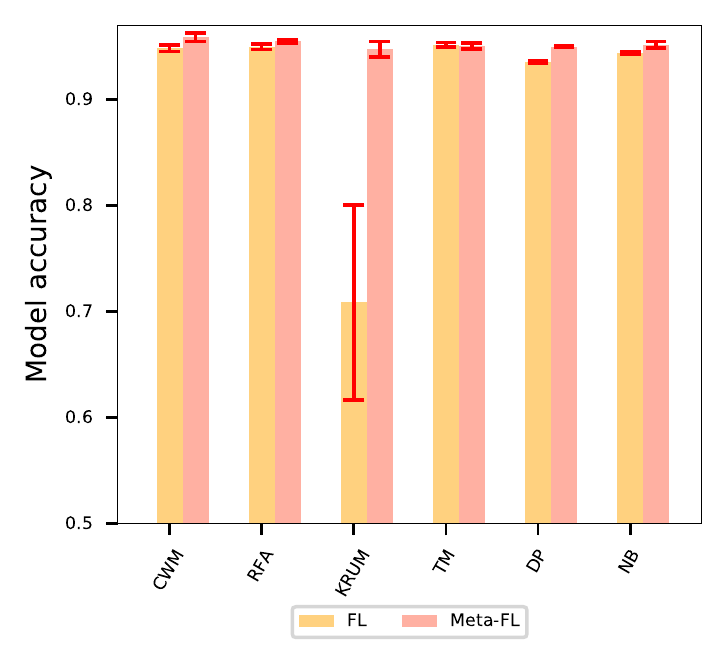}}%
\caption{Comparing utility of Meta-FL against baseline FL in terms of model accuracy. }
\label{utility-fig}
\end{figure*}

For ease of analysis, we assume that training cohorts are sampled independently meaning that there is no inter-dependency among client selection in each cohort. In this case, at any round $t$, updates submitted by any cohort of $c$ clients is essentially a random sample of size $c$ collected without replacement from the population of model updates. Assuming that updates are averaged as in Equation \ref{FedAvg-Eq}, cohort aggregates are in fact \emph{sample means} of model update population. As composition of cohorts is a random process, cohort aggregates are thus random variables whose distribution is determined by that of model updates as shown below (for proof refer to \citep{rice2006mathematical}) 
\begin{equation}
    \label{SRS-eq}
    Var(\Delta_{j})=\frac{\sigma_{j} ^{2}}{c} \left(\frac{P-c}{P-1} \right),\; \; \; \E[\Delta_{j}]=\E[\mu_{j}]
\end{equation}
Here, $\sigma_{j} ^{2}$ and $\mu_{j}$ denote variance and mean of population of model updates across the $j_{th}$ coordinate, respectively, and $\Delta_{j}$ indicates the $j_{th}$ coordinate of a cohort aggregate $\Delta$.
Assuming that each cohort contains more than one client ($1 < c$), it'd be trivial to show that $\frac{P-c}{c(P-1)}<1$. Therefore, we can prove that variance of cohort aggregates across any coordinate $j$ is upper bounded by variance of population of model updates across that coordinate as shown below:  
\begin{equation}
    \label{upper-bound}
     Var(\Delta_{j})=\sigma_{j} ^{2}\left(\frac{P-c}{c(P-1)} \right) < \sigma_{j} ^{2}
\end{equation}
A closer look at Equation \ref{upper-bound} reveals that server can further reduce the variance of cohort aggregates by increasing the size of training cohorts which would cause $\frac{P-c}{c(P-1)}$ becomes smaller and and closer to zero. Lower variation from cohort aggregates makes it easier for outlier detection based defenses to infer pattern of the benign observations, and effectively detect out-of-distribution malicious instances.

\textbf{Advantage 3.} As adversarial updates are aggregated with other updates, sybils face competition from benign clients to control value of cohort aggregate. This property makes it harder for the adversary to meticulously arrange values of adversarial aggregates to evade deployed defenses.

Our empirical evaluations in Section \ref{experiment-sec} will demonstrate that the advantages mentioned above in fact aid contemporary defense to perform better against backdoor attacks in Meta-FL compared to the baseline federated learning.

\begin{figure*}[t!]%
\centering
\subfigure[CWM]{%
\includegraphics[width=3.2cm,height=6cm]{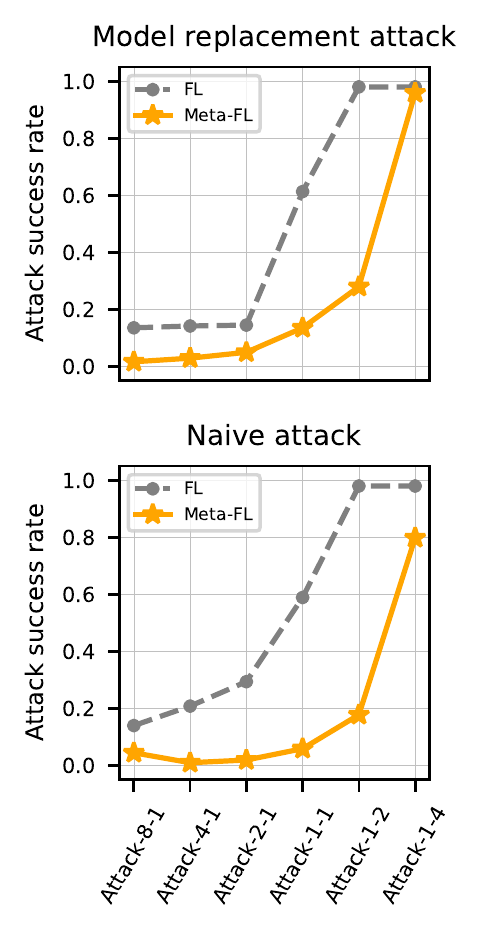}}%
\subfigure[Krum]{%
\includegraphics[width=2.8cm,height=6cm]{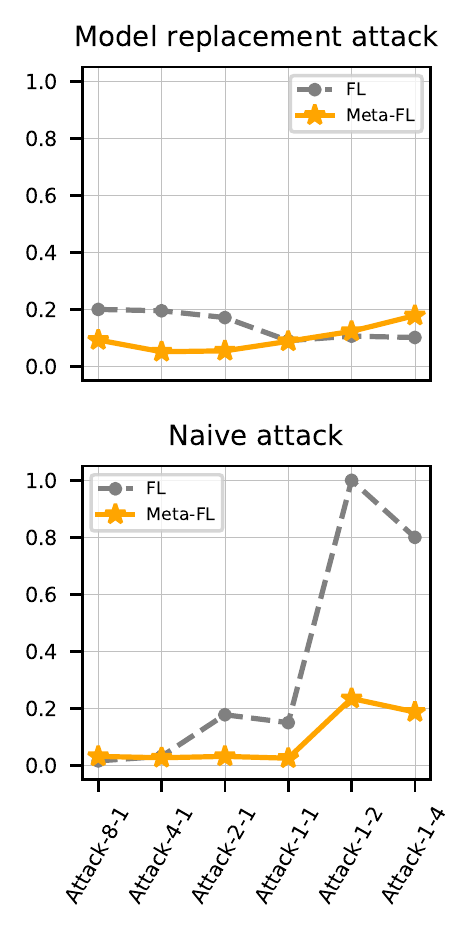}}%
\subfigure[Trimmed Mean]{%
\includegraphics[width=2.8cm,height=6cm]{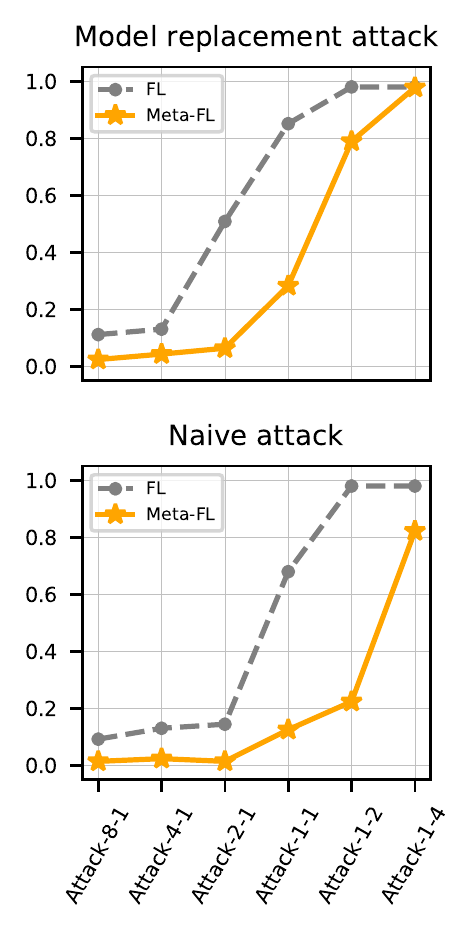}}%
\subfigure[RFA]{%
\includegraphics[width=2.8cm,height=6cm]{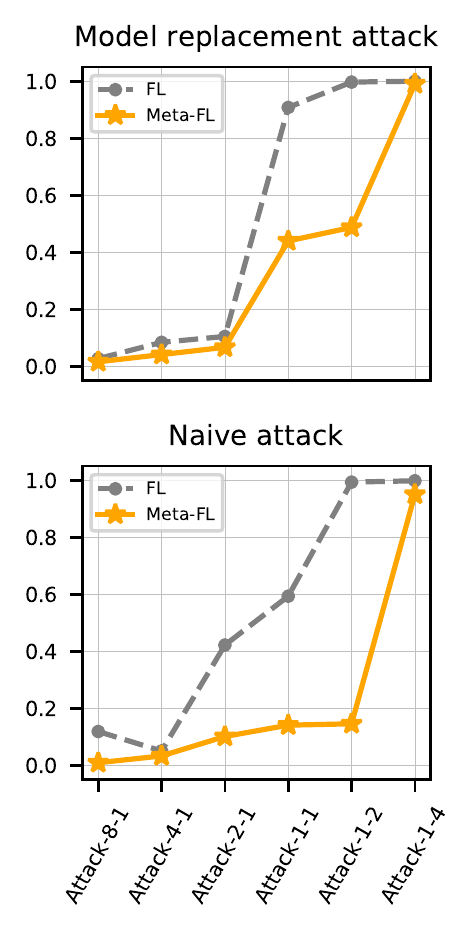}}%
\subfigure[Norm Bounding]{%
\includegraphics[width=2.8cm,height=6cm]{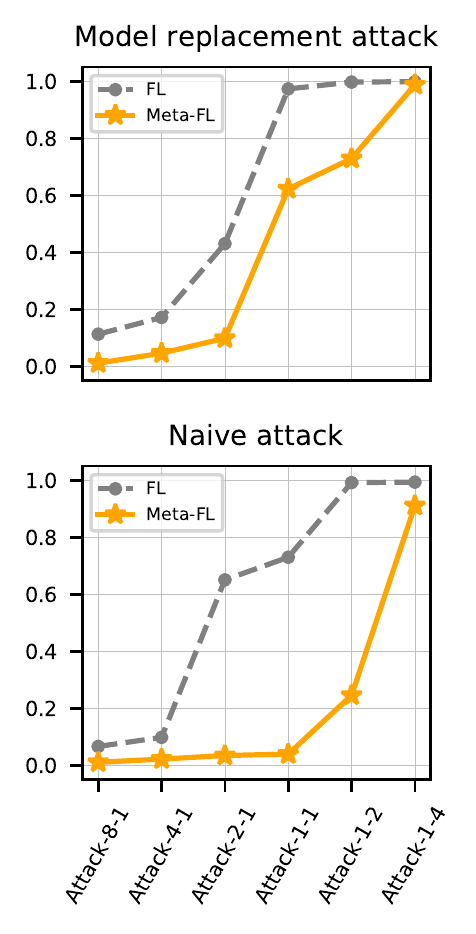}}%
\subfigure[Diff. Privacy]{%
\includegraphics[width=2.8cm,height=6cm]{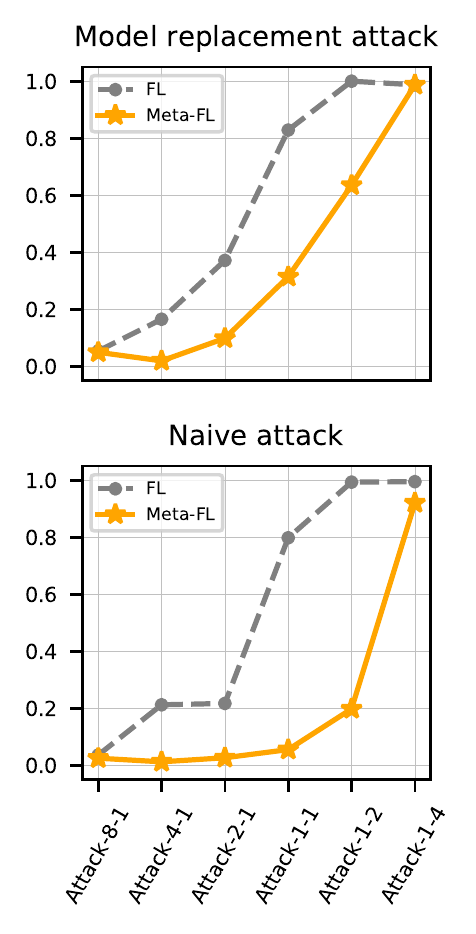}}%
\caption{Evaluating performance of contemporary defenses against naive and model replacement backdoor attacks on GTSRB model.}
\label{gtsrb-comp-fig}
\end{figure*}

\begin{figure*}[h!]%
\centering
\subfigure[CWM]{%
\includegraphics[width=3.2cm,height=6cm]{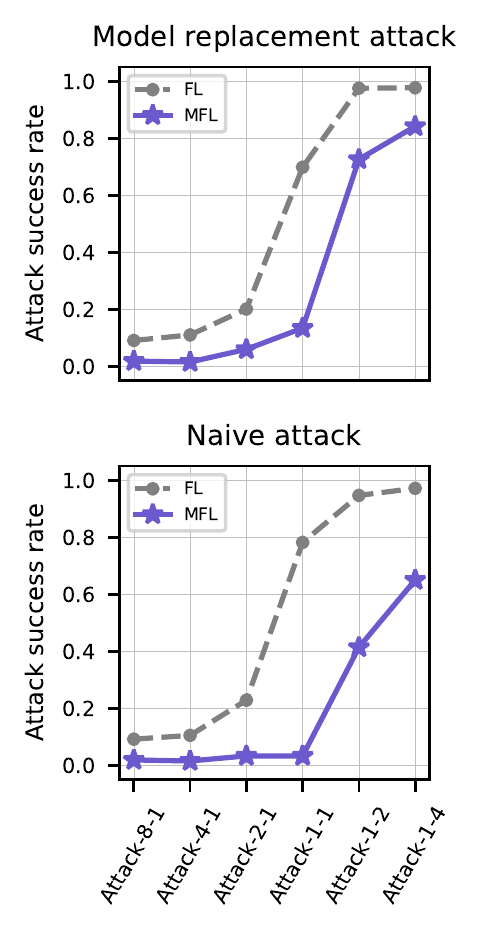}}%
\subfigure[Krum]{%
\includegraphics[width=2.8cm,height=6cm]{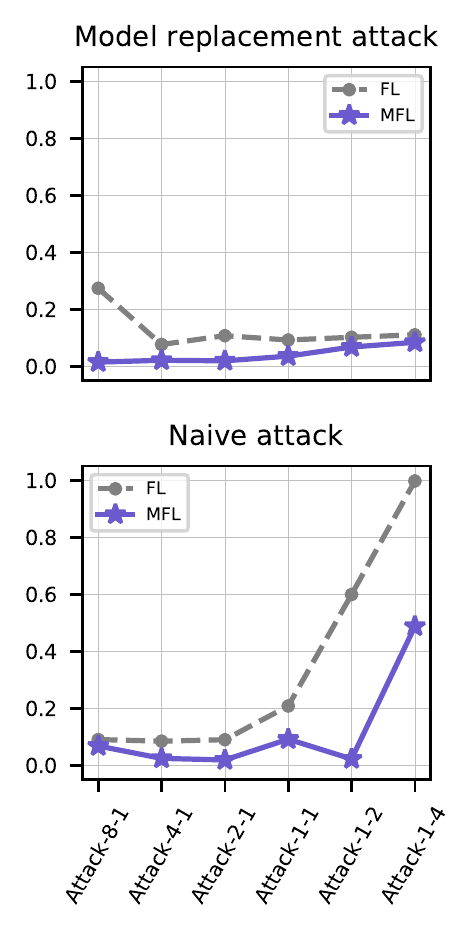}}%
\subfigure[Trimmed Mean]{%
\includegraphics[width=2.8cm,height=6cm]{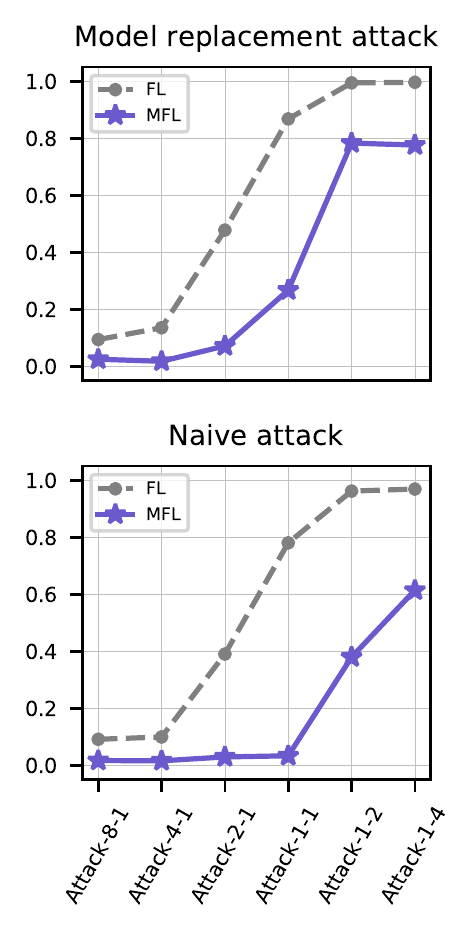}}%
\subfigure[RFA]{%
\includegraphics[width=2.8cm,height=6cm]{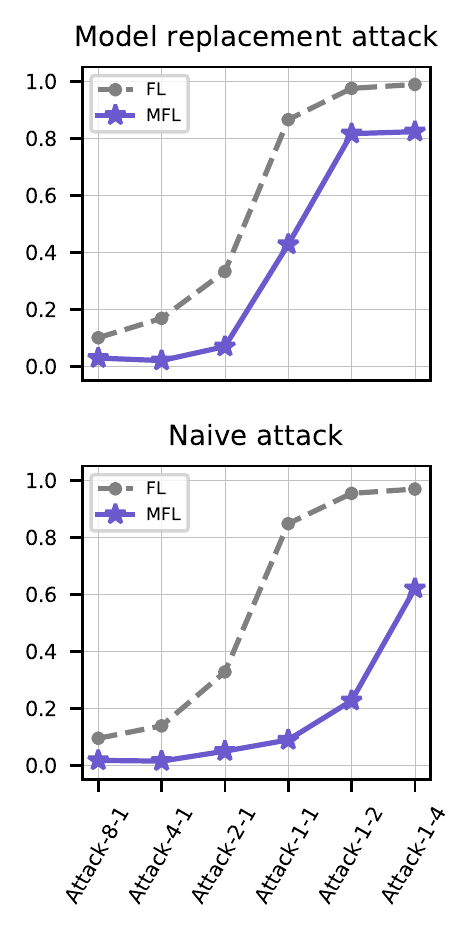}}%
\subfigure[Norm Bounding]{%
\includegraphics[width=2.8cm,height=6cm]{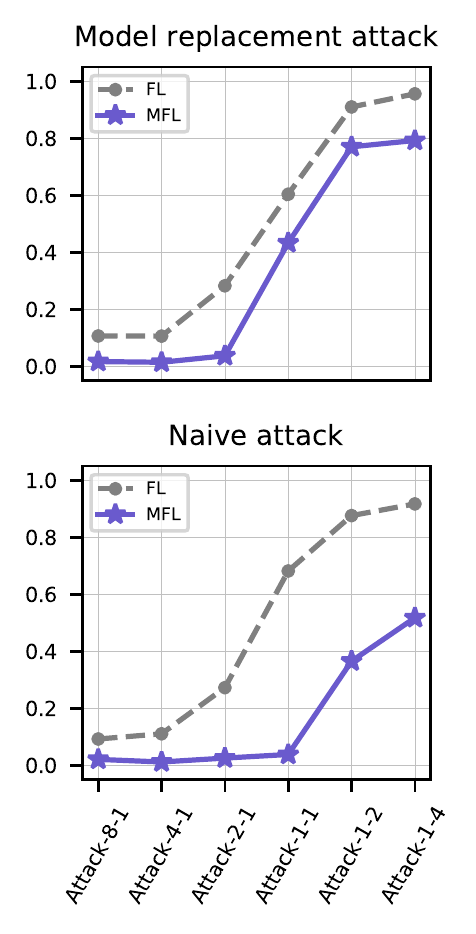}}%
\subfigure[Diff. Privacy]{%
\includegraphics[width=2.8cm,height=6cm]{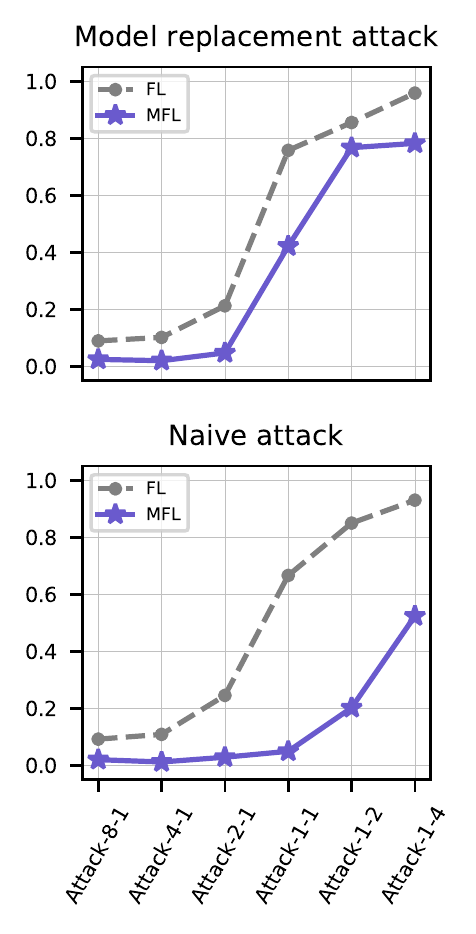}}%
\caption{Evaluating performance of contemporary defenses against naive and model replacement backdoor attacks on SVHN model.}
\label{svhn-comp-fig}
\end{figure*}

\section{Experiments}
\label{experiment-sec}

\subsection{Datasets and Experiment Setup}
We study Meta-FL on two classification datasets namely SVHN \cite{SVHN} and GTSRB \citep{GTSRB} with non-i.i.d. data distributions. \textbf{GTSRB} is s traffic sign dataset with 39,209 training and 12,630 test samples, where each sample is labeled with one of the 43 classes, and \textbf{SVHN} is a dataset of more than 100k images of digits cropped out of images of house and street numbers. For details on architecture and hyper parameters of benchmark models used for each dataset please refer to Appendix.

We use a Dirichlet distribution with parameter $\alpha=0.9$ to partition GTSRB and SVHN datasets into disjoint non-i.i.d shards and then distribute them among 150 and 300 clients, respectively. 
Following a similar set up to prior arts, each participating client trains their local model using SGD for 5 epochs with a batch size of 64 and learning rate of 0.1. Both Meta-FL and baseline FL resume the training process until certain number of training rounds are completed. Throughout our experiments, GTSRB and SVHN models are trained for 75 and 50 rounds, respectively.

For all experiments, pixel pattern backdoor attacks are performed in which adversary aims to influence model to misclassify inputs from a base label as a target label upon presence of an attacker chosen pattern (trigger). We set the adversarial trigger as a white square located at the top left corner of the image which roughly covers $9\%$ of the entire image. Objective of backdoor attacks in GTSRB and SVHN datasets are to mis-predict images of "Speed limit 80 miles per hour" as "Speed limit 50 miles per hour" and images of "digit 6" as "digit 1", upon presence of the white box trigger.

In the rest of paper, we denote each Meta-FL framework by two parameter as \textbf{MFL-i-j}, where $i$ and $j$ indicate number and size of training cohorts, respectively. We also use a similar notation \textbf{FL-k} to describe baseline FL systems, where $k$ indicates size of training cohort. 

Similar to the analysis in \citep{sun2019can}, we consider fixed frequency attack models to explore wide range of attack scenarios. In the baseline FL, \textbf{Attack-f-k} describes a scenario where $k$ sybils appear at every $f$ rounds of training to mount their attack. For the case of Meta-FL setting, \textbf{Attack-f-k} describes the case where at every $f$ round of training, $k$ training cohorts contain an adversarial client.

\subsection{Utility of Meta-FL}
In this section, we compare utility of Meta-FL against baseline setting in term of model accuracy. In this experiment, we evaluate utility of baseline and Meta-FL frameworks across various FL configurations, and aggregation rules. Note that for a fair comparison, we make sure number of clients participating in each round of model training are equal across both frameworks. Figure \ref{utility-fig} reports the test accuracy of models trained in Meta-FL and baseline settings deploying different defenses and aggregation rules. As reflected, federated training with Meta-FL results in more accurate models compared to baseline setting. All defenses and aggregation rules offer better utility in our framework. Even Krum aggregation rules which has been known
to cause a large drop in performance of learned model in baseline FL \citep{bagdasaryan2020backdoor,bhagoji2019analyzing} can train models with comparable performances in Meta-FL.

\begin{figure*}[t]%
\centering
\subfigure[SVHN]{%
\includegraphics[width=8.4cm,height=6cm]{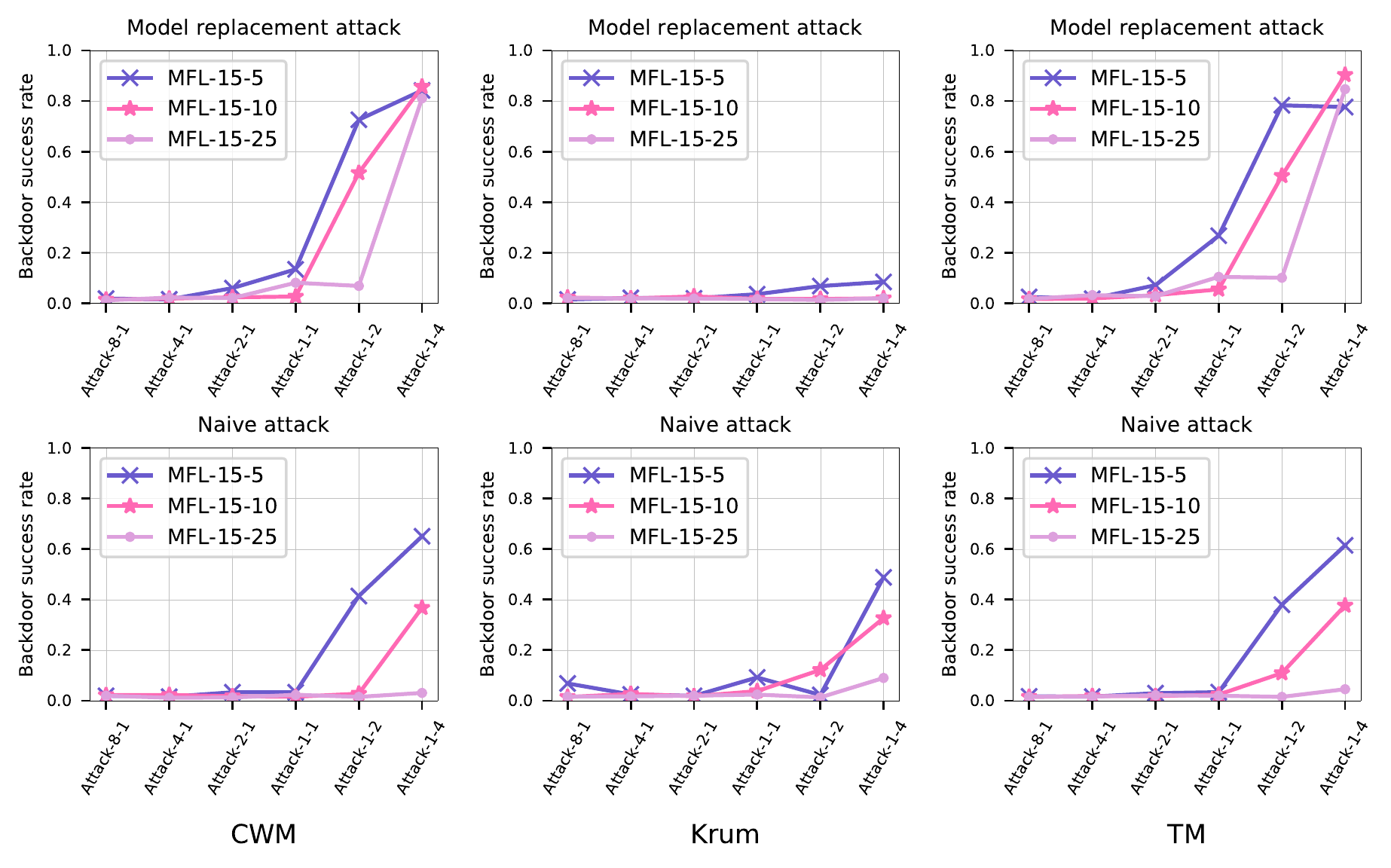}}%
\subfigure[GTSRB]{%
\includegraphics[width=8.4cm,height=6cm]{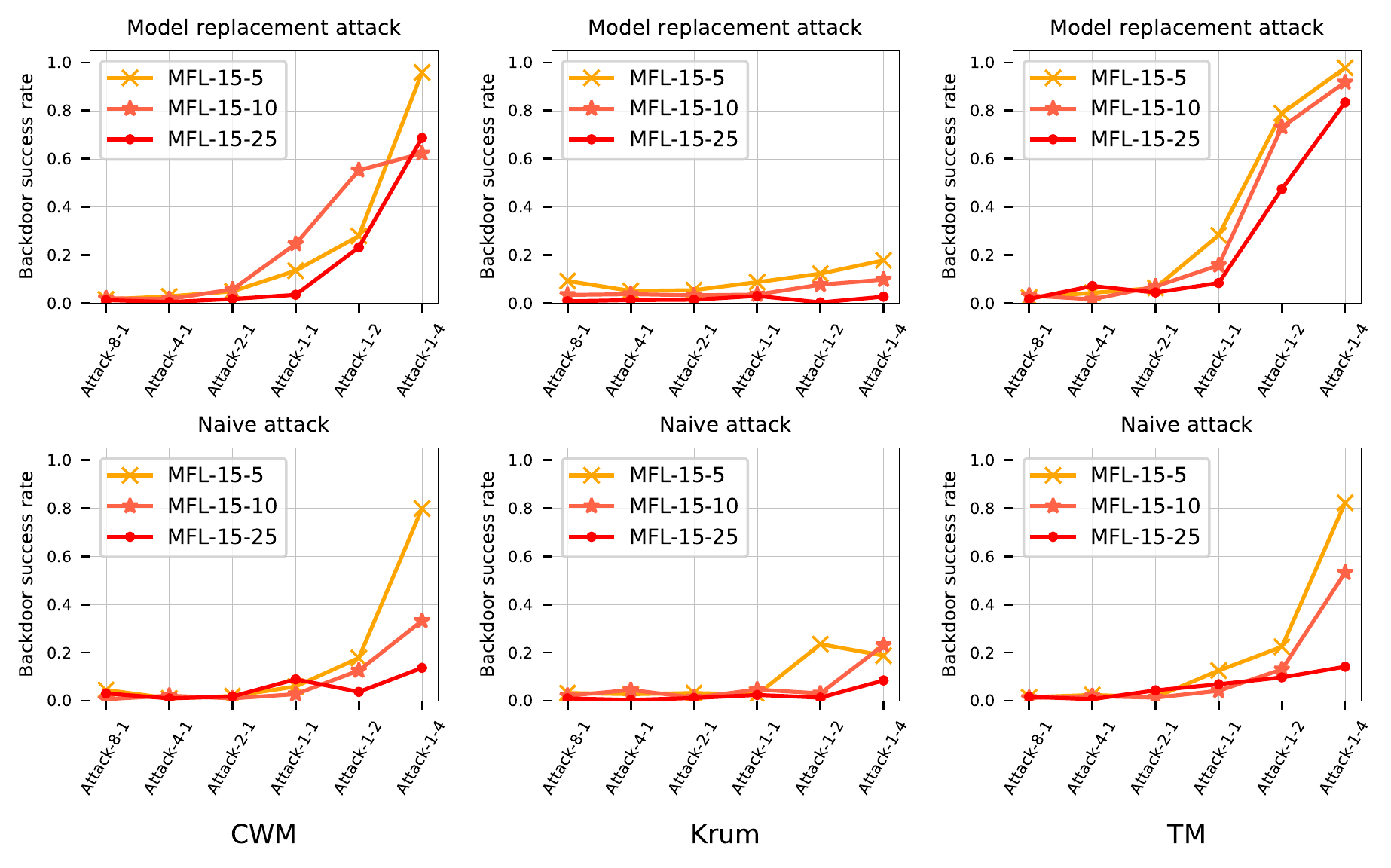}}%
\caption{Effect of size of training cohorts on efficacy of CWM, Krum and TM against backdoor attacks.}
\label{cohort-size-fig}
\end{figure*}

\subsection{Robustness of Meta-FL}
In this section, we systematically compare capabilities of contemporary defenses against backdoor attacks in both baseline and meta federated learning. Our empirical evaluation in this section shows that all defenses benefit from the advantages discussed in Section \ref{meta-fl-sec} and offer better robustness in our framework Meta-FL.

Figure \ref{gtsrb-comp-fig} and \ref{svhn-comp-fig} report performance of contemporary defenses against backdoor attacks on GTSRB and SVHN benchmarks, respectively.
We extend our experiments to both Meta-FL (MFL-15-5) and baseline FL (FL-5) frameworks for each dataset. In our evaluations, we consider defenses such as Krum, Coordinate-Wise Median (CWM), trimmed mean (TM), norm bounding (NB), differential privacy (DP) and RFA. For implementation details of these techniques please refer to Appendix.

We experiment with several attack scenarios to systematically evaluate performance of each defense against adversaries with wide range of resources at hand. As we move along the attack scenarios denoted on on horizontal axis of diagrams in Figures \ref{gtsrb-comp-fig} and \ref{svhn-comp-fig}, the adversary becomes more and more powerful, and appears more frequently with more sybils at each round.

For a fair evaluation of contemporary defense across Meta-FL and baseline FL, we make sure the defender faces similar challenges in both frameworks. Throughout our experiments in this section, we set number of training cohorts in Meta-FL equal to number of selected clients in baseline FL to ensure that server sees same number of "aggregands" (client updates in baseline FL and cohort aggregates in Meta-FL) across both cases. Moreover, the way our attack scenarios are defined ensures that same number of aggregands are adversarial across both framework.

Across both Meta-FL and baseline FL frameworks, the scaling factor for model replacement attack is set equal to the size of training cohorts to ensure that submissions from adversarial clients survive the averaging procedure and overpower aggregate of their corresponding cohort. For attack scenarios in which multiple sybils appear in the same round, we assume they coordinate and divide the scaling factor among themselves evenly.

Figure \ref{gtsrb-comp-fig} and \ref{svhn-comp-fig} shows that \emph{Meta-FL puts all defense at an advantage in mitigating against backdoor attacks.} Attack success rate of both the naive and model replacement approach in Meta-FL (solid lines) is lower than in baseline FL (dashed lines) when the same defense is in place across both frameworks. Therefore, our empirical evaluations shows that existing defenses are more robust to backdoor attacks in Meta-FL compared to baseline FL across.

While Meta-FL enhances robustness of all 6 methods, we observe that Krum benefits the most from our framework. We believe that lower variance on cohort aggregates aids Krum to effectively separate benign and malicious updates. We note that server can further decrease variance of cohort aggregates along each coordinate by increasing size of training cohorts, As discussed in Section \ref{meta-fl-sec}, and improve robustness of Krum aggregation rule.

Moreover, other methods such as coordinate-wise median and trimmed mean which are anomaly detection based defenses can also benefit from lower variations on cohort aggregate. Perhaps the most important principle in detecting outliers is defining the distribution of ordinary observations, which can be easier should observations exhibit low variations.
Figure \ref{cohort-size-fig} shows the results for experiments in which we evaluate performance of Krum, CWM and TM across Meta-FL frameworks with increasingly larger training cohorts. For this experiment, we set the number of cohorts to 15 and varied cohort size between 5, 10 and 15. As reflected in Figure \ref{cohort-size-fig}, increasing size of training cohorts improves robustness of these techniques across all scenarios, especially for scenarios in which adversary appears more frequently with more sybils.

Although defenses such as RFA, differential privacy and norm bounding appear to be robust against poisoning attacks \citep{sun2019can,pillutla2019robust}, our empirical evaluations shows that they are not effective against backdoor attacks, specifically model replacement attacks. In poisoning attacks, the adversarial sub-task, which is misclassification of unmodified data samples (e.g. classifying certain images of digit 1 as digit 7), is in direct contradiction with the primary learning task. Therefore, poisoning updates (or aggregates) face direct opposition from submissions of benign clients, which makes it harder for the adversary to succeed. However, for the case of backdoor attacks, adversary's goal for the model is to learn the causal relation between presence of an attacker chosen trigger and certain model output which does not require model to learn any knowledge contradicting the primary learning task. Therefore, backdoor attacks tend to be stealthier compared to poisoning attacks and defenses which have shown resilience against poisoning attacks might fall short against backdoor attacks.

\section{Conclusion}
In this paper, we show that it is in fact possible to defend against backdoor attacks without violating privacy of participating clients. We propose Meta-FL, a new federated learning framework which not only protects privacy of participants through the secure aggregation protocol but also facilitate defense against backdoor attacks. Our empirical evaluations demonstrate that state of the art defense tend to be more effective against backdoor attacks in Meta-FL compared to baseline FL while offering the same or better utility. Our results suggest that not only does Meta-FL protect privacy of participants but also optimizes the robustness-utility trade off better than baseline setting.
\bibliography{example_paper}
\bibliographystyle{mlsys2020}

\clearpage
\appendix

\section{Implementation details of defenses and attacks}
As mentioned in Section \ref{experiment-sec}, we compare performance of contemporary defense such as Krum, Coordinate-wise Median (CWM), trimmed mean (TM), norm bounding (NB), differential privacy (DP), and RFA against naive and model replacement backdoor attacks on both Meta-FL (\textbf{MFL-15-10}) and baseline federated learning ($\textbf{FL-15}$) frameworks in which number of aggregands $n$ across both frameworks are equal.
Hyper parameter and implementation details of these techniques are provided below: \textbf{(1)} For Krum, to meet the convergence condition $n \geq 2f + 3$, we set $f=6$.
\textbf{(2)} In Trimmed mean, the parameter $\beta$ is set to $0.20$.
\textbf{(3)} For RFA, the maximum iteration of Weiszfeld algorithm and the smoothing factor is set to 10, and $10^{-6}$, respectively.
\textbf{(4)} In norm bounding defense, as the original work \citep{sun2019can} did not provide a recipe to decide the norm threshold $M$, we developed our own approach to determine $M$. In our experiments, at each round, we set the norm threshold $M$ to the norm of smallest aggregand to ensure all aggregands will have equal $l2$ norm before aggregation. In federated learning, as the global model converges, model updates (and therefore cohort aggregates) start to fade out and have smaller norms. Therefore, setting a constant norm threshold for all training rounds would not be effective, which is why we took a dynamic approach to decide $M$.   
\textbf{(5)} For differential privacy the hyper-parameter $M$ is set similar to norm bounding and then a Gaussian noise $\mathcal{N}(0.0,\,0.001^{2})$ is added to aggregate of updates (or cohort aggregates) before updating the global model.

\section{Architecture and Hyperparameters of Benchmark Models}
\label{model-archs}
Table \ref{modelarch} reports the topology and hyper parameters of benchmarks used for GTSRB and SVHN datasets. 
\begin{table}[h]
  \caption{Model architecture for SVHN and GTSRB datasets.}
  \label{modelarch}
  \centering
  \begin{tabular}{cccc}
    \hline
    \multicolumn{2}{c}{GTSRB}  & \multicolumn{2}{c}{SVHN}                 \\
    \cmidrule(r){1-2}  
    \cmidrule(r){3-4}
    Layer Type     & Filter/Unit     & Layer Type     & Filter/Unit \\
    \midrule
    Conv + ReLU & $3 \times 3 \times 32$  & Conv + ReLU & $3 \times 3 \times 32$    \\
    Conv + ReLU & $3 \times 3 \times 32$  & Conv + ReLU & $3 \times 3 \times 32$    \\
    Conv + ReLU & $3 \times 3 \times 64$  & Conv + ReLU & $3 \times 3 \times 64$    \\
    Conv + ReLU & $3 \times 3 \times 64$  & Conv + ReLU & $3 \times 3 \times 64$    \\
    Conv + ReLU & $3 \times 3 \times 128$  & Conv + ReLU & $3 \times 3 \times 128$    \\
    Conv + ReLU & $3 \times 3 \times 128$  & Conv + ReLU & $3 \times 3 \times 128$    \\
    FC + ReLU & 43  &  FC + ReLU & 512  \\
    Softmax & 43 & FC + ReLU & 10 \\
    && Softmax & 10 \\
    \bottomrule
  \end{tabular}
\end{table}

\end{document}